%% file: ms.tex
\documentclass[twoside]{article}

\usepackage[accepted]{aistats2023}
\usepackage{hyperref}       
\usepackage{url}            
\usepackage{booktabs}       
\usepackage{amsfonts}       
\usepackage{nicefrac}       
\usepackage{microtype}      
\usepackage{xcolor}         
\usepackage{graphicx}
\usepackage{subfigure}
\usepackage{multirow}
\usepackage{threeparttable}
\usepackage{makecell}
\usepackage{color}
\usepackage{nicefrac}
\usepackage{booktabs} 
\usepackage{nicematrix}

\usepackage{amsmath, amssymb, amsthm, mathtools} 
\usepackage{wrapfig}  
\usepackage{mdwlist}  
\usepackage{sidecap}  
\usepackage{bbm}
\usepackage{bm,amsbsy} 

\usepackage{caption}
\usepackage{algorithm}
\usepackage{algpseudocode}
\usepackage{fullpage}
\usepackage{cancel} 
\usepackage{ulem} 

\usepackage{stmaryrd}
\usepackage{MnSymbol}

\usepackage{newfloat}
\usepackage{listings}
\DeclareCaptionStyle{ruled}{labelfont=normalfont,labelsep=colon,strut=off} 
\floatstyle{ruled}
\newfloat{listing}{tb}{lst}{}
\floatname{listing}{Listing}

\input{notation.tex}

\begin{document}

%

%

\twocolumn[

\aistatstitle{Differential Privacy Meets Neural Network Pruning}

\aistatsauthor{ Kamil Adamczewski \And Mijung Park}

\aistatsaddress{ ETH Z\"urich, MPI for Intelligent Systems \And  University of British Columbia} ]

\begin{abstract}
A major challenge in applying differential privacy to training deep neural network models is scalability.
The widely-used training algorithm,  \textit{differentially private stochastic gradient descent
(DP-SGD)}, struggles with training moderately-sized neural network
models for a value of $\varepsilon$  corresponding to a high level of privacy protection. 
In this paper, we explore the idea of dimensionality reduction inspired by neural network pruning to improve the scalability of DP-SGD. 
We study the interplay between neural network pruning and differential privacy, 
through the two modes of parameter updates. We call the first mode, \textit{parameter freezing}, where we pre-prune the network and only update the remaining parameters using DP-SGD. We call the second mode, \textit{parameter selection}, where we select which parameters to update at each step of training and update only those selected using DP-SGD.
In these modes, we use public data for freezing or selecting parameters to avoid privacy loss incurring in these steps. Naturally, the closeness between the private and public data plays an important role in the success of this paradigm. 
Our experimental results demonstrate how decreasing the parameter space improves differentially private training. Moreover, by studying two popular forms of pruning which do not rely on gradients and do not incur an additional privacy loss, we show that random selection performs on par with magnitude-based selection when it comes to DP-SGD training.


\end{abstract}

\section{Introduction}

Differential privacy (DP) \cite{Dwork14} is a gold standard privacy notion that is widely used in many applications in machine learning.
Generally speaking, the \textit{accuracy} of a model trained with a dataset trades off with the guaranteed level of \textit{privacy} of the individuals in the dataset. 
%
One of the most popular differentially private (DP) training algorithms for deep learning is \textit{DP-SGD} \cite{DP_SGD}, which adds a carefully calibrated amount of noise to the gradients during training using stochastic gradient descent (SGD) to provide a certain level of differential privacy guarantee for the trained model at the end of the training. Since it adds noise to the gradients in every training step, its natural consequence is that the accuracy of the model drops.  
Typically, the trade-off between the privacy and accuracy gets worse  when the model size gets larger, as the dimension of the gradients increases accordingly.

In the current state of differentially private classifier training, the performance between small and large models is striking. A small two-layer convolutional neural network trained on MNIST data with $\epsilon=2$ provides a classification accuracy of $95\%$ \cite{Da21}. However, a larger and more complex architecture, the ResNet-50 model, containing $50$-layers, trained on ImageNet data with $\epsilon < 5$ provides a mere classification accuracy of $2.4\%$, while the same model without privacy guarantee achieves $75\%$ accuracy \cite{jax_dp}. 
Given that this is the current-state-of-the-art in the literature, we cannot help but notice a disappointingly huge gap between the differeitally privately trained and non-privately trained classifiers.

Why is there such a large gap between what the  current state-of-the-art deep learning can do and what the  current state-of-the-art differentially private deep learning can do? The reason is that those deep models that can reliably classify high-dimensional images are extremely \textit{large-scale} and DP-SGD does not scale to these large models. The definition of DP requires that every parameter dimension has to be equally guarded. For example, consider two models, where first model has $10$ parameters and second model has $1000$ parameters, where these parameters are normalized by some constant (say $1$) to have a limited sensitivity, which is required for the DP guarantee. As we need to add an equal amount of independent noise to each of the  parameters to guarantee DP, where the parameters are normalized by the same constant, the signal-to-noise ratio per parameter becomes roughly $100$ times worse in the second model with $1000$ parameters compared to the first model with $10$ parameters. From this observation, it might look almost dispiriting to make any reasonably sized deep neural network models differentially private and maintain their original accuracies.

In this work, inspired neural network pruning, we explore the impact of reducing the parameter space on differentially private training. Respectively, we apply pruning in two forms, (1) parameter freezing where the set of trainable network parameters is decreased prior to the private training, and (2) parameter selection, where the parameter space is preserved but only a subset of weights is updated at every iteration.   

This work results in a number of intriguing observations:
\begin{enumerate}
    \item For training in a differentially private fashion, pruning in both forms improves the network performance at a given privacy level, compared to DP-SGD. Interestingly, we found that our pruning-based dimensionality reduction is as effective as the PCA-based dimensionality reduction like that in \cite{Da21}, while these two use an entirely different criterion to reduce the dimensionality of the parameters. 
    \item At small $\epsilon$ (i.e., in high privacy regime), we found that higher pruning rates improves the accuracy of the DP trained models. 
    \item Parameter freezing improves the performance more at small $\epsilon$ (i.e., in high privacy regime), while parameter selection fares better in medium-to-low privacy levels. 
    \item Surprisingly, random selection fares the same or better as selecting weights based on the magnitude of the gradients. Similar observations have been made in \cite{random_select}, but not in the context of transferring knowledge from public to private data.
    \item Similar to transfer learning, the more similar private and public data are, the better the performance of our algorithm gets even in the presence of noise for privacy. 
\end{enumerate}

Experimental results 
prove to narrow the gap between the performance of privately trained models and  that of non-privately trained models by improving the privacy-accuracy trade-offs in training large-scale classifiers.
In what comes next, we start by describing relevant background information before introducing our method. 


\section{Background}
This work builds on the ideas from neural network pruning. While neural network pruning has a broad range of research outcomes, we summarize a few relevant works that help understand the core ideas of our algorithm. We then also briefly describe differential privacy for an unfamiliar audience.  

\subsection{Neural Network Pruning}
\label{subsec:pruning}

The main reasoning behind neural network pruning is that neural networks are typically considered over-parametrized, and numerous observations indicate that pruning out non-essential parameters does not sacrifice  a model's performance by much. There are two types of pruning techniques: one-shot pruning and iterative pruning. 

\textbf{One-shot pruning.} One-shot pruning is the process that follows a three-step procedure. The network is trained from scratch. This model, also known as the pre-trained model, is the base for network pruning. Then given some criteria, the model is pruned, and some parameters are removed which results in a new architecture which is a sub-architecture of the original model. This step, however, results in the loss of accuracy. To compensate for this loss, the remaining network is fine-tuned to recover the accuracy.
This is the idea we will use in our first mode, parameter freezing, where we pre-train a model with a public dataset and pre-prune the parameters based on a certain criterion. We then train these remaining parameters using DP-SGD.

\textbf{Iterative pruning.} The second mode is iterative pruning. Iterative pruning is similar to one-shot pruning but the three-step process known from one-shot pruning is repeated multiple times and the parameters are pruned gradually. While in one-shot pruning, the parameter removal happens at one time, iterative pruning proceeds over several iterations. At each iteration, a fraction of parameters is removed and the rest of the network is fine-tuned. 
This is the very idea we will use in our second mode, parameter selection. We select parameters to update in every training step, where this selection is done based on public data, and update the selected ones using private data with DP-SGD.


Consequently, we adapt these ideas to improve differentially private training 
by reducing the space of the parameter to be updated. This distinguishes our work from the majority of the  neural network compression literature, which aims to compress the model. 
 This procedure is supposed to alleviate the accuracy-privacy trade-off and shift the Pareto line. Subsequently, we present the core privacy ideas which stand behind the above-mentioned trade-off. 




\subsection{Differential privacy and DP-SGD}

A mechanism $\mathcal{M}$ is  ($\epsilon$, $\delta$)-DP, if the output of the mechanism given a dataset $\Dat$ is \textit{similar} to that given a neighbouring dataset $\Dat'$ where neighbouring means $\Dat$ and $\Dat'$ have one sample difference, and the similarity is defined by a selected value of $\epsilon$ and $\delta$, which quantify the privacy guarantees. 
Mathematically, when the following equation holds, we say $\mathcal{M}$ is  ($\epsilon$, $\delta$)-DP: 
$
\Pr[\mathcal{M}(\Dat) \in S] \leq e^\epsilon \cdot \Pr[\mathcal{M}(\Dat') \in S] + \delta,
$
 where the inequality holds for all possible sets of the mechanism's outputs $S$ and all neighbouring datasets $\Dat$, $\Dat'$.
%

The \textit{Gaussian mechanism} is one of the most widely used DP mechanisms. As the name suggest, the Gaussian mechanism adds noise drawn from a Gaussian distribution, where the noise scale is proportional to two quantities, the first one is, so called, {\it{global sensitivity}} (which we denote by $\Delta$) and the second one is a privacy parameter (which we denote by $\sigma$). 
The global sensitivity of a function $g$
\cite{dwork2006our} denoted by $\Delta_g$ is defined by the maximum difference in terms of $L_2$-norm $||g(\Dat)-g(\Dat') ||_2$, for neighbouring $\Dat$ and $\Dat'$. In the next section we will describe the relationship between global sensitivity and gradient clipping in case of neural networks.
The privacy parameter $\sigma$ is a function of $\epsilon$ and $\delta$. Given a selected value of $\epsilon$ and $\delta$, one can readily compute $\sigma$ using numerical methods, e.g., 
the auto-dp package by \cite{wang2019subsampled}.

It is worth noting two  properties of differential privacy.  The \textit{post-processing invariance} property \cite{dwork2006our} states that the composition of any data-independent mapping with an $(\epsilon,\delta)$-DP algorithm is also $(\epsilon,\delta)$-DP. 
%
%
The
\textit{composability} property \cite{dwork2006our} states that the strength of privacy guarantee degrades in a measurable way with repeated use of the training data. 
%

In the neural network training, we first take a mini-batch of data samples drawn from the training dataset, then compute the sum or average of the gradients evaluated on the mini-batch.  Here, we simply do not know how much one datapoint's difference would change the L2 norm of the sum or the average of the gradients. Hence, we adopt a simple  clipping procedure to ensure the average or sum of the gradients has a limited sensitivity.   
Once we clip the sample-wise gradient -- the gradient evaluated on each datapoint in the mini-batch -- we add the Gaussian noise, where the noise scale is proportional to the clipping norm and $\sigma$ in case of sum, or the clipping norm and $\sigma$ divided by the mini-batch size in case of average. The resulting algorithm is called DP-SGD\cite{DP_SGD}.
Due to the composability of differential privacy, the cumulative privacy loss increases as we increase the number of training epochs (i.e., as we access the training data more often with a higher number of epochs). When models are large-scale, the required number of epochs until convergence increases as well, which imposes more challenges in achieving a good level of accuracy at a reasonable level of privacy. Our method, which we introduce next,  provides a solution to tackle this challenge. 

\section{Method}

We propose a differentially private training framework that uses public data for pre-training a large-scale classification model and then fine-tunes the model to private data by updating only a few selected gradients using DP-SGD. We call our method, \textit{differentially private \textbf{sparse} stochastic gradient descent} (DP-SSGD). 
The summary of our algorithm 
is given in \figref{algo}.  

\subsection{Vanilla DP-SGD}

Suppose a neural network model with an arbitrary architecture denoted by $f$, and parameterized by $\theta$. We first describe the vanilla DP-SGD algorithm \cite{DP_SGD}. As in the regular stochastic gradient descent, at each training step $t$, a dataset mini-batch $\mathcal{B}^{(t)}$ (where $|\mathcal{B}|=B$) is drawn, and, given that mini-batch, the gradient of the loss $l$  is computed. However, to ensure the privacy, the learning needs to be controlled, and thus the $L2$-norm of the gradients is bounded with the, so-called, clip-norm $C$. As such, the gradient is given by  
$g = \frac{1}{B} \sum_{b \in \mathcal{B}^{(t)}} \nabla^C[b]$, where the sample-wise gradient $\nabla^C[b] = \frac{\partial}{\partial \theta} l(b)$ is norm-clipped such that $\|\frac{\partial}{\partial \theta} l(b) \|_2 \leq C$ for all $b \in \mathcal{B}^{(t)}$ to ensure the limited sensitivity of the gradients. We then add noise to the average gradient such that
\begin{align}
    \theta^{(t+1)} \; & \leftmapsto \;  \theta^{(t)} - \frac{\eta }{B} \left[ \sum_{b \in B^{(t)}} \nabla^C[b] + \sigma C \zeta \right]
\end{align} where $\eta$ is the learning rate, and $\sigma$ is the privacy parameter, which is a function of a desired DP level $(\epsilon, \delta)$ (larger value of $\sigma$ indicates higher level of privacy), and $\zeta$ is the noise such that $\zeta \sim \Nrm(0, I)$. Note that higher clipping norms $C$ result in higher levels of noise added to the gradient.  How our method deviates from DP-SGD is in the following way.

\subsection{Sparsification}
\label{sec:sparsification}
The key tool used for our method is gradient pruning which selects which weight parameters are updated. For a given layer, let $I$ be a list of vectors such that $i \in I$ and $i \in \mathbb{R}^4$ for convolutional layers where the indices of $i$ describe output, input and two dimensional kernel dimensions, and $i \in \mathbb{R}^2$ for fully-connected layers, where the indices describe output and input dimensions. As a result of this selection step, we split the index set $I$ of all the weights into two groups. The first group is an index set $I_{s}$ that contains all the selected indices for the gradients that we will update in the following step; another index set $I_{ns}$ that contains all the non-selected (or discarded) indices for the gradients that we will  not update. Moreover, define pruning rate $p$ as the fraction of the gradients which are discarded. Then $I = I_s \cup I_{ns}$

\textbf{Parameter index selection.}
Now at time $t$, we subsample a a mini-batch $\mathcal{B}^{(t)}$ from the private dataset and compute the gradients on the private mini-batch.
As in DP-SGD, we clip the sample-wise gradients. However, for each $b \in \mathcal{B}$, we clip \textit{only} the selected weights indicated by the index set $I_s$:

\begin{align}
    \nabla^{C} &= \frac{\nabla^{C}[I_s, b]}{ \max\left(1, \frac{|\nabla^{C}[I_s, b]|}{C}\right)},
\end{align} 
Smaller gradient norm benefits gradients in two crucial ways. Firstly notice that decreasing the size of the gradients will be clipped only if $|\nabla^{C}[I_s, b]| > C$. Moreover, they are clipped proportionally to the value $\frac{|\nabla^{C}[I_s, b]|}{C}$, therefore small gradient norm results in less clipping. Reducing the parameter size results in decreasing the norm and preserving larger gradient values. In other words, we trade off small gradients for keeping the large signal. 

Whether we use the full-size gradients or the gradients of selected gradients, if we use the same clipping norm $C$, the sensitivity of the sum of the gradients in both cases is simply $C$.  
However, adding noise to a vector of longer length (when we do not select weights) has a worse signal-to-noise ratio than adding noise to a vector of shorter length (when we do select weights).

Moreover since the neural network models are typically over-parameterized and contain a high level of redundancy, we observe that dropping a large portion of gradients does not hamper the training. This means the length of $I_s$ can be significantly smaller than the length of $I$. This is useful in reducing the \textit{effective} sensitivity of the sum of the gradients.

We then update the parameter values for those selected using DP-SGD:
\begin{align}
    \vtheta^{(t+1)}[I_s] &=\hat\vtheta^{(t)}[I_s] - \frac{\eta_t}{B} \left[\sum_{b \in \mathcal{B}^{(t)}}\nabla^{C}[I_s, b] + \sigma C \zeta \right],
\end{align} where $\zeta \sim \Nrm(0, I)$.
For the parameters corresponding to the non-selected gradients indicated by $I_{ns}$, we simply do not update the values for them, by   
\begin{align}
    \vtheta^{(t+1)}[I_{ns}] &=\vtheta^{(t)}[I_{ns}].
\end{align} We finally update the parameter values to the new values denoted by $\vtheta^{(t+1)}$. We repeat these steps until convergence.

\subsection{Two modes of sparsification}

In the previous section, we define the sparsification. In this section, we describe sparsity constraints, that is which weights can be updated at every iteration. 
We distinguish two modes of sparsifying the neural network training, inspired by neural network pruning. In the first mode given the pretrained model, we select a fixed set of parameter indices that are removed from the original network and the remaining parameters are fine-tuned. The second one is to maintain the original dimensions of the network and, at every iteration, select the subset of parameters to be updated. We describe in detail the two modes below.

\paragraph{Parameter freezing.} We first assume parameter redundancy. According to neural network pruning literature, a model can be compressed by removing redundant parameters, and retain the original performance via retraining. In this line of thinking, the index set of selected weights, $I_s$, remains fixed throughout the training (which coresponds to a compressed model in network pruning). That is for any iterations $t_i$ and $t_j$, $I_s^{t_i}=I_s^{t_j}$.

Let us note in this work we are not concerned about the model size, unlike in network pruning literature. That is as it is given in Sec. 3  we do not remove the parameters, per se, but we freeze them. That is, $p$ weights are frozen and the remaining weights are updated during the fine-tuning procedure.

\input{alg1.tex}

\paragraph{Parameter selection.} We now assume gradient redundancy. In one-shot pruning, the model is pruned prior to the retraining and then a fixed set of parameters is fine-tuned. 

Nevertheless, our task is not concerned with decreasing the size of the model, and hence removing weights may be unnecessarily restrictive. In our case, we allow each of the weights to be updated at every iteration, however only some of weights will be updated at a given iteration. That is for two iterations $t_i$ and $t_j$ generally, $I_{t_i} \neq I_{t_j}$ but $I_{t_i} \subset I_t$ and $I_{t_j} \subset I_t$.  

Notice that parameter freezing and variable parameter selection are two opposing modes. In the first mode, we allow no flexibility to the model and fix the parameter set at the beginning of the training. In the second mode, we allow for maximum flexibility as no parameter is discarded and all the parameters can be updated with the private data.  




\subsection{Pruning criteria}

In the previous section we introduce the constraints which parameters in the network can be updated. In this section, we describe exactly how the parameters are selected given those constraints. In this work we consider two criteria, (1) Random pruning and (2) Magnitude pruning.

\paragraph{Random pruning.} Given a pre-trained model, and pruning ratio $p \in [0,1)$, a set of $p$ weights for each layer is drawn uniformly at random and removed.

\paragraph{Magnitude pruning.} The importance of each individual weight is judged by its absolute value. To compress the network, we sort the parameters of each layer according to their magnitude. The bottom $p$ parameters are removed, and the remaining top $(1-p)$ are fine-tuned. 

While there is a host of pruning methods, not all can be applied for the privacy training provides additional constraint. The two criteria we suggest here have two main advantages. Firstly, they are both simple and have little computational overhead. Random pruning involves straightforward uniform sampling. Magnitude pruning is the most popular pruning benchmark whose selection is based on sorted weights. The second advantage is that they do not incur additional privacy cost. Random selection only involves selecting indices without looking at data samples. Magnitude pruning is done based on the current model parameters. This model has already been privatized and using it does not cause any further privacy violation.

Finally, let us note the significant difference between pruning a network and the proposed DP-SSGD. In pruning, both the pre-training and fine-tuning/retraining are performed on the same training dataset. In our work, we distinguish two datasets, public and private. We assume the public dataset is freely available and can be accessed without sacrificing the privacy of data. We can then exclude it in the privacy analysis. The second dataset consists of private data whose access should be restricted and privatized. Ideally, the public dataset should be similar to the private dataset. More precisely, both public and private datasets should contain high-level features that are similar and can be interchangeably used in the trained model. 

\section{Related work}

Recently, there has been a flurry of efforts  on improving the performance of DP-SGD in training large-scale neural network models. These efforts are made from a wide range of angles, as described next. 
For instance, 
\cite{lyu2021dpsignsgd} developed a sign-based DP-SGD approach called DP-SignSGD for more efficient and robust training.
\cite{Papernot_Thakurta_Song_Chien_Erlingsson_2021} replaced the ReLU activations with the tempered sigmoid activation to reduce the bias caused by gradient clipping in DP-SGD and improved its performance. 
%
%
Furthermore, 
\cite{shamsabadi2022losing}
suggested modifying the loss function to promote a fast convergence in DP-SGD. 
\cite{DPNAS} developed a paradigm for 
neural network architecture search with differential privacy constraints, while
\cite{wang2019differentially} a gradient hard thresholding framework that provides good utility guarantees. 
%
%
\cite{10.1145/3469877.3490594} proposed a grouped gradient clipping mechanism to modulate the gradient weights.
\cite{scale_norm} suggested a simple architectural modification termed ScaleNorm by which an additional normalization layer is added to further improve the performance of DP-SGD. 

The transfer learning idea we use in our work also has been a popular way to save the privacy budget in DP classification. 
For instance, 
 \cite{tramer2021differentially} suggested transferring features learned from public data to classify private data with a minimal privacy cost. \cite{CVPR_Sparse} developed an architecture which consists of an encoder and a joint classifier, where these are pre-trained with public data and then fine-tuned with private data. When fine-tuning these models, \cite{CVPR_Sparse} also used the idea of neural network pruning to sparsify the gradients to update using DP-SGD.    
Our method is more general than that in  \cite{CVPR_Sparse} in that we sparsify the gradients given any architecture, while \cite{CVPR_Sparse} focuses on the particular architecture they introduced and specific fine-tuning techniques for the architecture.



Many attempts which take advantage of public data have improved the baseline performance. A recent work improved the accuracy of a privately trained classifier (ResNet-20) for CIFAR10 from $37\%$ to $60\%$ accuracy at $\epsilon=2$ \cite{Da21}.
Another work improved the accuracy of a privately trained classifier (ResNet-18) for ImageNet from $6\%$ to $48\%$ accuracy at $\epsilon=10$ \cite{jax_dp}. However, there is still a huge room to improve further, given that the non-privately trained classifiers of the same architecture for each of these datasets reach $90\%, 75\%$, respectively. 


Existing works for reducing the dimension of gradients in DP-SGD also use public data for estimating the subspaces with which the dimension of the gradients are reduced \cite{kairouz2021fast, zhou2020bypassing, Da21, pmlr-v139-yu21f}. \cite{random_select} suggested random selection of gradients to reduce the dimensionality of the gradients to privatize in DP-SGD. 
Unlike these works, we also reduce the dimension of the gradients by selecting the gradients on the magnitude of each weight or weight, inspired by neural network pruning. Besides, \cite{random_select} proposed to freeze the parameters progressively. We introduce two different (and complementary) settings where the fixed set of parameters is frozen, and where no parameters are frozen. 



\section{Experiments}

\input{plots1.tex}

We conduct experiments on two datasets, MNIST and CIFAR-10. Following \cite{Da21}, MNIST is trained on a small convolutional neural network with two convolutional and one fully-connected layer, CIFAR-10 is trained on Resnet-20. 
We replace all batch normalization layers with group normalization as group normalization averages only within a sample, not across samples (like batch normalization) which prevents proper privacy analysis. 

The networks are first pre-trained with public datasets. 
In case the number of classes of the pre-trained dataset is different, the last layer is removed and randomly initialized. 
Then,  
the networks are trained on a private dataset at three privacy levels, $\epsilon=2, 5$ and $8$. The smaller number indicates higher levels of privacy and $\infty$ stands for non-private training. The parameter details are given in the Appendix.


\subsection{DP-SSGD pruning comparison}

In the DP-SSGD pruning comparison study we compare two modes, parameter freezing and parameter selection. The former decreases the size of the model prior to the training with private data with the pruning ration $p$, and the latter maintains the original size and varies a subset of updated weights over the course of the training such that at every iteration $p$ gradients are discarded.

In both modes, random and maximum magnitude criteria are used. Then each of the four settings is tested on three levels of private training to test the behavior of the proposed method on different privacy levels, strong, medium and weak. Moreover, we vary the pruning level $p$ from 0 (no pruning) to 0.9. Every set-up is tested five times and we report the average across them.

For CIFAR-10, we treat CIFAR-100 as public data and pre-train Resnet-20 on CIFAR-100 (accuracy 6.8\% on CIFAR-10). For MNIST, we pre-trained the small convolutional network with a $0.05\%$ subset of MNIST (96.1\% on MNIST), which we treat as public data. The study on the interplay between public and private data is given in 

Figure ~\ref{fig:dpssgd} describes the summary of the results. The findings are absorbing. For small $\epsilon$ increasing the pruning rate actually improves the model performance. There is a positive correlation between the pruning rate and the test accuracy. This is the case both for parameter freezing and variable selection. That is, the network learns more with fewer gradients updated in the presence of high levels of noise (further confirming the claim that the signal-to-noise is increased as the pruning increases.). This is in contrast to the training with lower levels of privacy where pruning the gradients hurts the accuracy. The medium level of privacy (middle column in Fig. ~\ref{fig:dpssgd}) is the case in-between and indeed shows ambiguous results. Pruning may be beneficial up to certain degree after which point, the test accuracy decreases. Further hyperparameter tuning may be beneficial to determine optimal pruning ratio for a given privacy setting.

The relationships between pruning and privacy level are more pronounced for larger Resnet architectures than for smaller convolutional networks. As a result, larger networks may benefit from the DP-SGDD training where the gradients are pruned. But when trained in non-private setting, pruning gradients is likely to hurt its performance.

Intuitively, using all the weights increases the norm of the parameter matrix, which triggers the clipping of weights. On the one hand, clipping affects the larger weights more than the smaller ones, thus decreasing the values of the gradients which make the most difference. On the other hand, smaller weights impact learning negatively in twofold ways. As given by the pruning literature, they are less relevant for model learning, and secondly, in a private setting, they cause the larger, more relevant weights to lose their information by decreasing their gradient value. 

Finally, what may be seen as surprising is that magnitude pruning performs similarly to random pruning, which may mark another difference between pruning the private and non-private training. In non-private setting, magnitude pruning is often proved effective as a network compression mechanism.

\subsection{Benchmark comparison}

\input{table.tex}

The numerical results are summarized in Table \ref{tab:table1}. The results are presented for three datasets, CIFAR, MNIST and Imagenet. In the proposed method, DP-SSGD, we present two types of results, parameter freezing and variable parameter selection, along with DP-SGD and GEP \cite{Da21}, which also studies gradient redundancy and produces low-dimensional gradient embedding. The numbers come from the original paper.

The results are consistent with the ablation study. 
Parameter freezing performs better for higher levels of privacy (and smaller levels of epsilon), and variable parameter selection gains an edge when the training gets closer to the non-private setting. 
In the parameter freezing setting, we optimize over a fixed, smaller set of parameters which, it can be hypothesized, increases their signal-to-noise over the course of the training. On the other hand, the larger parameter space in the case of variable selection may result in the noise dominating the learned signal. 


\subsection{Which public data to use?}

So far, we showed the results with a fixed public dataset. 
We use public data to pre-train the model before the private training starts, and also to select which parameters to update in the parameter selection mode based on the magnitude. So, the proximity  between the public and private dataset pairs plays an important role in the success of our algorithm. An intuitive reasoning is that if the public data is more similar to private data, one needs a less number of fine-tuning steps with differential privacy to achieve the same accuracy. We test this reasoning in our next experiment.

For differentially private CIFAR-10 classifier, we tested CIFAR-100, SVHN, and 5\% of CIFAR10 data as public data. The DP-trained classifiers' accuracies are summarized in Table \ref{tab:table2}.
\input{table_pub_priv}

\input{table_pub_priv_mnist}
For differentially private MNIST classifier, we tested SVHN and 5\% of MNIST data as public data. The DP-trained classifiers' accuracies are summarized in Table \ref{tab:table3}.

From these experiments, we conclude that for simple datasets like MNIST, using a small portion of the private data as public data provides a good initialization before the private training. However, for more complex data like CIFAR10, using the small portion of the same dataset does not provide good initialization. Rather, it is encouraged to look for datasets that are more complex than the private data as public data.  

\section{Conclusion}

In this paper, we conclude that, as far as the training of highly parametrized networks is concerned, differentially private training with limited privacy budget curbs the possibility to train all the parameters at once in the same way as the non-private training. Furthermore, this work shows the positive relationship between decreasing the updated parameter space and training performance. Given varying privacy budget, disparate modes of sparsifications may be preferred. Finally, random parameter selection proves to be a remarkable way of sparsifying network for differentially private training.



\bibliographystyle{plain}
\bibliography{bib}

\end{document}

%% file: notation.tex
\newcommand{\punt}[1]{}

\usepackage{amsthm}
\usepackage{xcolor}
\usepackage{hyperref}

\renewcommand{\eqref}[1]{eq.~\ref{eq:#1}}
\newcommand{\Nrm}{\mathcal{N}}

\newcommand{\figref}[1]{Fig.~\ref{fig:#1}}  

\newcommand{\Dat}{\mathcal{D}}

\newcommand{\vtheta}{\mathbf{\ensuremath{\bm{\theta}}}}

\newcommand{\vn}{\mathbf{n}}

\newcommand{\vg}{\mathbf{g}}

\def\Pr{\ensuremath{\text{Pr}}}

%% file: alg1.tex
\begin{figure}

\begin{algorithm}[H]
\caption{DP-SSGD}\label{algo:dp_grad_drop_var1}
\begin{algorithmic}[1]
\Require Tensor  $\nabla$ that contains the  gradients evaluated on each data sample in a mini-batch $S$, pruning rate $C$ (a constant) that determines the compression rate of the resulting average gradient matrix (average across data samples), norm clipping hyperparameters $\gamma$, criterium $\leftarrow \{max, rand\}$. 
\Ensure Sparse and differentially private average gradient matrix
\Function{Select-indices}{$C$, criterium, weights}  
    \State $I \leftarrow$ indices of weights 
    
    \If{Random}
        \State $I_{ns}$ $\leftarrow p \cdot len(I)$ indices of $I$, selected uniformly at random 
    \Else
        \If{Max}
            \State $I_{as} \leftarrow$ argsort(weights)
            \State $I_{ns} \leftarrow$ bottom $p \cdot len(I)$ indices of $I_{as}$ 
        \EndIf
    \EndIf
\State $I_s \leftarrow I \setminus I_{ns}$
\Return {$I_{ns}$}
\EndFunction
\State
\Function{DP-SGD}{$\nabla$, $I_{ns}$, $\gamma$, $\sigma$}
    \State We compute the average gradient matrix: $\vg = \frac{1}{B} \sum_{b=1}^S \nabla[:,s]$, where $b \in \mathcal{B}$.
   \State  We ensure  $  \nabla^2[I_s] \leq \gamma$ by clipping the squared norm by $\gamma$.
    \State We privatize the gradients in the selected index set,  $\nabla \leftarrow \nabla[I_s] + \vn$, where $\vn \sim \Nrm(0, \sigma^2 C^2 I)$.
    
\Return {$I_{ns}, I_s$}
\EndFunction
\State
\Function{DP-Sparse-SGD}{$\nabla,C, \sigma, \gamma, $ criterium, $\mathcal{B}$, weights}

    \If{freezing}
     \State $I_{ns} \leftarrow$ \Call{Select-indices}{}
    \EndIf

    \For{batch $\mathcal{B}$}        
    \If{selection}
     \State $I_{ns} \leftarrow$ \Call{Select-indices}{}
    \EndIf
    \State $I \leftarrow \nabla[I_{ns}] = 0$
    \State $\nabla_{DP-SSGD} \leftarrow$ \Call{DP-SGD}{}
    \EndFor
\Return{$\nabla_{DP-SSGD}$} 
\EndFunction
\end{algorithmic}
\end{algorithm}

    \caption{DP-SSGD algorithm. The algorithms consists of two parts, indices selection based on criterium and privatized gradient update. In case of freezing mode, indices selection is made only once before the training. In case of selection mode, indices are selected at every iteration. Notice we do not use any gradient information to select the indices, thus preserving privacy budget.}
    \label{fig:algo}
\end{figure}

%% file: plots1.tex
\begin{figure*}[h]
    \centering
    \begin{minipage}{.33\textwidth}
    \includegraphics[scale=0.33]{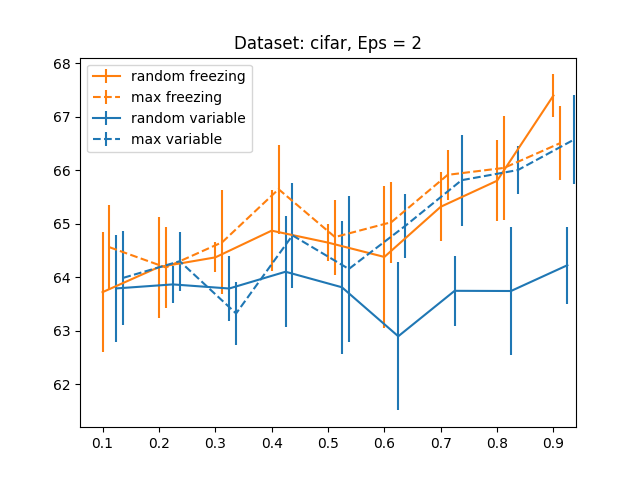}
    \end{minipage}\hfill
    \begin{minipage}{.33\textwidth}
        \includegraphics[scale=0.33]{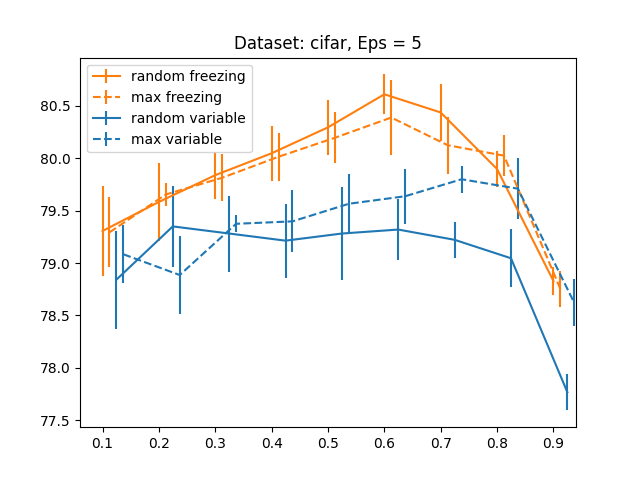}
    \end{minipage}\hfill
    \begin{minipage}{.33\textwidth}
        \includegraphics[scale=0.33]{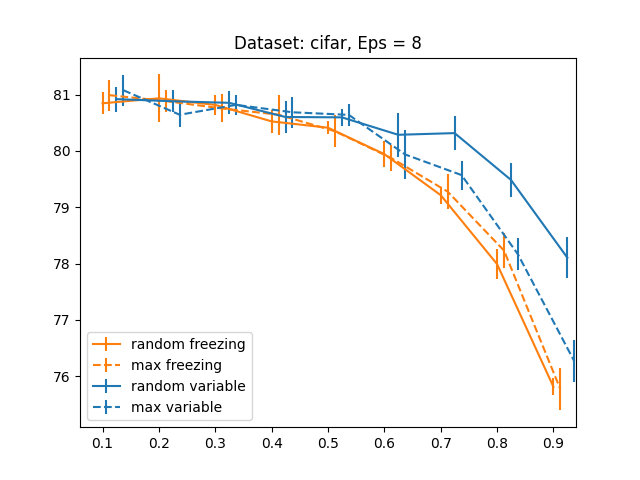}
    \end{minipage}\hfill
        \begin{minipage}{.33\textwidth}
    \includegraphics[scale=0.33]{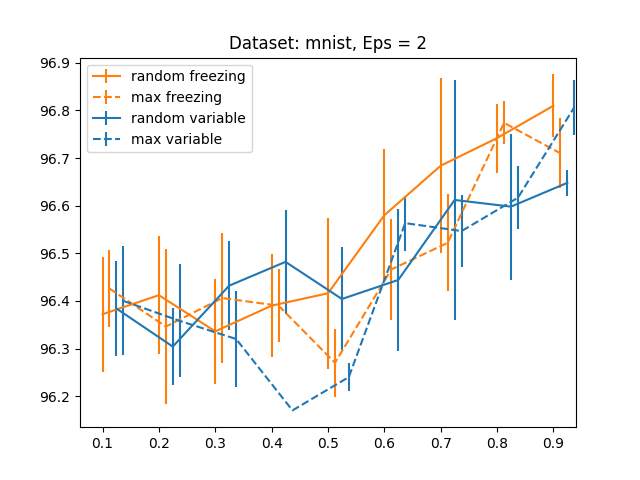}
    \end{minipage}\hfill
    \begin{minipage}{.33\textwidth}
        \includegraphics[scale=0.33]{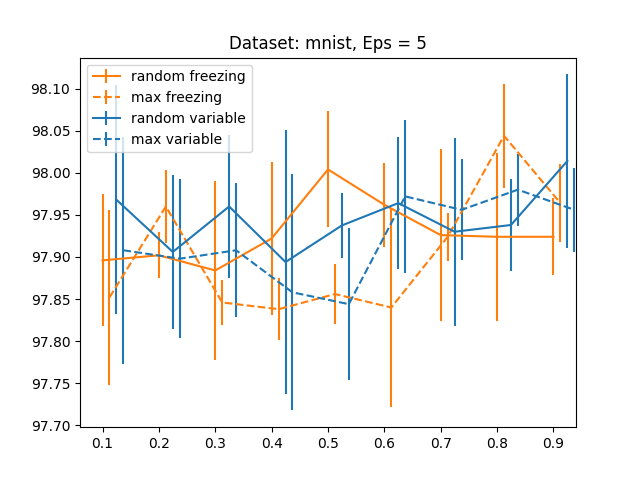}
    \end{minipage}\hfill
    \begin{minipage}{.33\textwidth}
        \includegraphics[scale=0.33]{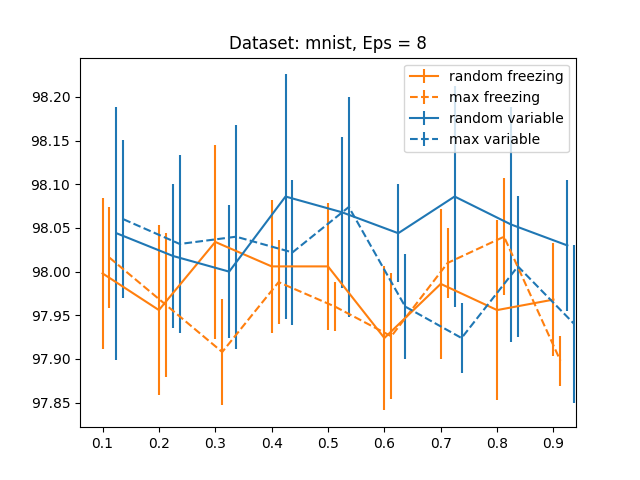}
    \end{minipage}\hfill
    \caption{The effect of pruning rate on the accuracy of the differentially private training for a fixed epsilon. The plots on the top depicts Resnet training on Cifar-10, at the bottom, small convolutional network trained on MNIST. In the figure legend, random freezing means randomly selected parameter freezing, max freezing means the parameters whose magnitude is large are frozen. Similarly, random variable means parameter selection uniformly at random during training; and max variable means parameter selection based on the magnitude during training. 
    }
    \label{fig:dpssgd}
    \vspace{0.3cm}
\end{figure*}

%% file: table.tex
\begin{table*}[ht!]
    \centering
\begin{tabular}{cccccc}
\toprule
CIFAR-10 & $\epsilon$ & Rand (param freezing) & Rand (param selection) & GEP & DP-SGD \\ \midrule
& 2 & \textbf{72.11} & 70.11& 59.7 & 68.5 \\
& 5 & \textbf{81.09 }& 79.99 & 70.1 & 79.8 \\
& 8 & 81.75 &\textbf{81.9} & 74.9 & 81.3 \\
 \midrule
MNIST & $\epsilon$ & Rand (param freezing) & Rand (param selection) & GEP & DP-SGD \\ \midrule
& 2 & \textbf{97.02 }& 96.96 & 96.3 & 97.5 \\
& 5 & 98.09 & \textbf{98.18} & 97.9 & 98.0 \\
& 8 & 98.17 & 98.26 & \textbf{98.4} & 98.05 \\ 
\bottomrule
\end{tabular} 
    \caption{Test accuracy for two datasets, CIFAR-10 and MNIST trained with differential privacy constraints. For DP-SSGD, we present the results at three privacy levels $\epsilon=2,5,8$. For DP-SSGD, we present the performance of random pruning. DP-SGD numbers take into account pre-training on the public model and are obtained via DP training without pruning}
    \label{tab:table1}
\end{table*}

%% file: table_pub_priv.tex
\begin{table}[ht!]
    \centering
\begin{tabular}{cccc}
\toprule
$\epsilon$ & CIFAR100  & SVHN  & CIFAR10 ($5\%$)  \\ \midrule
 2 & 72.11 & 11.87 & 54.35  \\
 5 & 81.09 & 18.56 & 53.9  \\
 8 & 81.9 & - & 62.13 \\
\bottomrule
\end{tabular} 
    \caption{For training a classsifier for CIFAR-10, we tested three different public datasets. CIFAR100, which is similar to CIFAR-10, SVHN, which is quite different from SVHN, and 5\% of CIFAR-10, which is the same as the private dataset while the quantity is not sufficient enough to provide a good classification performance. These numbers are from varying pruning rates and sparsification modes.}
    \label{tab:table2}
\end{table}

%% file: table_pub_priv_mnist.tex
\begin{table}[ht!]
    \centering
\begin{tabular}{ccc}
\toprule
$\epsilon$ & SVHN  & MNIST ($5\%$)  \\ \midrule
 2 & 95.14 & 97.02  \\
 5 &  96.47 & 98.18  \\
 8 &  96.84 & 98.26 \\
\bottomrule
\end{tabular} 
    \caption{For training a classsifier for MNIST, we tested two different public datasets. SVHN, which is more complex than MNIST, and 5\% of MNIST. Unlike the CIFAR-10 data, for MNIST, the pre-trained classifier with only 5\% of MNIST data provides a good initialization for private training. These numbers are from varying pruning rates and sparsification modes.}
    \label{tab:table3}
\end{table}